\documentclass[sigconf]{acmart}

\usepackage{tikz}
\usetikzlibrary{shapes,arrows}
\usetikzlibrary{automata,positioning,fit,backgrounds}
\usetikzlibrary{arrows.meta}

\AtBeginDocument{%
  }

\setcopyright{acmlicensed}
\copyrightyear{2024}
\acmYear{2024}
\acmDOI{XXXXXXX.XXXXXXX}

\acmConference[KDD 2024 Workshop]{Causal Inference and Machine Learning in Practice}{August 26, 2024}{Barcelona, Spain}
\acmISBN{978-1-4503-XXXX-X/18/06}

\begin{document}

\title{Industrial-Grade Smart Troubleshooting through Causal Technical Language Processing: a Proof of Concept}

\author{Alexandre Trilla}
\affiliation{%
  \institution{Alstom, Santa Perp\`etua de la Mogoda\\
	La Salle, Universitat Ramon Llull}
  \city{Barcelona}
  \country{Spain}}
\email{alexandre.trilla@alstomgroup.com}

\author{Ossee Yiboe}
\affiliation{%
  \institution{Alstom, Saint Ouen}
  \city{Paris}
  \country{France}}
\email{ossee-josepha-charlesia.yiboe@alstomgroup.com}

\author{Nenad Mijatovic}
\affiliation{%
  \institution{Alstom, Saint Ouen}
  \city{Paris}
  \country{France}}
\email{nenad.mijatovic@alstomgroup.com}

\author{Jordi Vitri\`a}
\affiliation{%
  \institution{Universitat de Barcelona}
  \city{Barcelona}
  \country{Spain}}
\email{jordi.vitria@ub.edu}

\renewcommand{\shortauthors}{Trilla et al.}

\begin{abstract}
This paper describes the development of a causal diagnosis
approach for troubleshooting an industrial environment on the 
basis of the technical language expressed in Return on Experience 
records. The proposed method leverages the vectorized linguistic knowledge 
contained in the distributed representation of a Large Language Model, 
and the causal associations entailed by the embedded failure modes 
and mechanisms of the industrial assets.
The paper presents the elementary but essential concepts of the 
solution, which is conceived as a causality-aware retrieval augmented 
generation system, and illustrates them experimentally on a real-world 
Predictive Maintenance setting. Finally, it discusses avenues of improvement
for the maturity of the utilized causal technology to meet the robustness
challenges of increasingly complex scenarios in the industry.
\end{abstract}

\maketitle

\section{Introduction}
\label{secIntro}
The degradation of industrial assets is a complex multifaceted problem 
that can be explained by different factors. As the components
wear and deteriorate, the systems exhibit a series of changes that
increase in severity until they eventually fail. In this case,
failure patterns may also emerge. For instance, in the
Reliability Engineering field, assets are most expected to fail
either prematurely (early) during their break-in period, or late by
the end of their remaining useful life (wear-out)~\cite{Dersin23}.
These failure types can be \emph{anticipated} because their modes, 
mechanisms, and effects, are well known and documented.
In consequence, engineers introduce
quality checks in the manufacturing process and inspection actions 
in their (more-or-less conservative) preventive maintenance schedule
to mitigate their impact. However, for as long as the machines 
operate, failures may seem to appear ``randomly'' at any point in 
time. This is especially challenging for
dependable assets while they transit the middle risky region, when 
the failure rate is relatively low, but uniform/constant.

In this uncertain setting, the field of Predictive Maintenance
tackles the problem by introducing the data as a means to closely
follow the actual evolution of each asset and make better
informed and timely decisions~\cite{Fink20}.
In this sense, the detection of incipient anomalous
behaviors and the capacity to \emph{diagnose} their 
\emph{root causes} and \emph{predict} their \emph{solutions}
towards a more favorable \emph{prognosis} become
increasingly important to guarantee the availability of the
machines. 

To succeed in these multiple objectives, the required information 
and knowledge, which displays a clear \emph{causal} character, 
is typically described and compiled in textual form through
two different (linguistic) environments~\cite{Bunge08}. 
On the one hand, an \emph{ontological}
reference framework based on a Failure Mode, Mechanism,
and Effect Analysis (FMMEA)~\cite{iec60812},
which provides a scholarly
structure of causality driven by degradation. On the other hand,
a \emph{methodological/epistemological} 
approach via an actual record on Return 
On Experience (RoX), the data of which have been explicitly 
written for the purpose of explaining both the root causes and solutions
of the reported failures. In both environments, several experts 
inherently identify which properties of
the observations describe spurious correlations unrelated to the
causal explanation of interest, and which properties represent
the phenomenon of interest, i.e., the stable invariant associations.

Traditional approaches for processing language in Predictive Maintenance
settings have initially considered the
idiosyncrasies of technical environments~\cite{Brundage21}, and
have evolved into exploiting Large Language Models~\cite{Lukens23},
ontologies~\cite{Woods24}, and extracting recurrent problems and 
frequently suggested solutions~\cite{Sala24}.
Almost concurrently, the community of natural language processing
and computational linguistics identified causal challenges in textual
data~\cite{Feder21}, and these were soon also considered in the 
technical domains as a means to explain the
degradation mechanisms by developing custom word
embeddings~\cite{Trilla22TLP} and unconfounded subsystem
structures~\cite{Valcamonico24}.

This workshop paper holds the hypothesis that root causes and solutions
can be learned directly from the textual expressions found in
field-specific RoX data, which are fully aligned with the Smart 
Troubleshooting objective: given the description of a problem, the system
shall be able to accurately determine the related root cause and solution.
The paper is organized as follows: Section~\ref{secBackground}
reviews the modeling fundamentals of Causality and
Language, Section~\ref{secMethod} describes the
proposed causal diagnosis method from the standpoint of standard
Predictive Maintenance, Section~\ref{secResults} illustrates
the implementation of the method on a specific illustrative 
example, Section~\ref{secDisc} discusses avenues of improvement
to increase the robustness of the approach, and
Section~\ref{secConc} concludes the work.

\section{Background}
\label{secBackground}

This section presents the basic notions on how to model
the environment under analysis, both from a causal and a
linguistic perspective.

\subsection{Structural Causal Model}
\label{subSCM}

The causal links among the variables $X$ that build the model of
a system are assumed to be most effectively represented using the 
tools from the field of Causality. In this sense, the Structural Causal 
Model (SCM) is the framework that can most generally capture such 
directed associations~\cite{Pearl19}. The SCM defines a set of assignments
governing their specific functional associations $f$, along with 
some independent exogenous noise $N$ that accounts for everything 
that is not explicitly included in the model:
\begin{equation}
\label{eqSCM}
	X_j := f_j(PA_j, N_j)\ ,
\end{equation}
where $PA_j$ represents the direct causes of the $X_j$ variable.

If enough knowledge and experience from the field is available
from the subject matter experts, e.g., through the FMMEA or RoX
structures, then a complete SCM graph may be developed right 
from the start. Otherwise, the data need to be 
carefully leveraged to drive the discovery of the causal model.

\subsubsection{Causal Bayesian Network}
\label{subsubCBN}

Once the structural graph that binds the variables is determined, the 
functional associations of the SCM may be learned, 
and this work specifically adopts a
stochastic interpretation of the world. Therefore, it treats
all $X$ as random variables, and the resulting SCM statistically 
describes their (conditional) probability distributions.

Considering $n$ random variables $X_1,X_2,...,X_n$ and a
directed acyclic graph that relates them causally, a Causal
Bayesian Network (CBN) is a generative model that has the following
factorized joint probability distribution:
\begin{equation}
\label{eqBN}
	P(X_1,...,X_n) = \prod_{j=1}^{n} P \left ( X_j | PA_j, N_j \right ) \ .
\end{equation}

The graphical nature of Bayesian networks allows seeing relationships 
among different variables, and their conditional dependencies
enable performing probabilistic inference~\cite{Alaeddini11}.
In particular, CBN are powerful tools for knowledge representation 
and inference under \emph{uncertainty}~\cite{Pourret08}.

\subsection{Causal Inference}
\label{subCI}

Beyond probabilistic inference, Causal Inference provides the tools 
that allow estimating causal conclusions from observational data,
i.e., in the absence of a true experiment, given that certain 
assumptions are fulfilled. These 
assumptions increase in strength as is defined in Pearl's Causal 
Hierarchy (PCH) abstraction~\cite{Bareinboim22}, which is
summarized as follows for the purposes of this paper.

\subsubsection{PCH Rung 1: Associational}
\label{subsubPCH1}
Describes the observational distribution of the factual data through
their joint probability function $P(X)$. From this point forward,
interesting quantities, i.e., the queries $X_Q$, can be directly 
computed given some evidence $X_E$ through their conditional
probability, which is computed as a ratio of marginals:
\begin{equation}
\label{eqQE}
	P(X_Q | X_E) = \frac{P(X_Q,X_E)}{P(X_E)} \ .
\end{equation}

This level of analysis displays a degree sophistication akin to 
classical (un)supervised Machine Learning techniques. As such, it is
subject to \emph{confounding bias}, where common causes may induce
spurious statistical associations/correlations~\cite{Reichenbach56}.

\subsubsection{PCH Rung 2: Interventional}
\label{subsubPCH2}
Describes an actionable distribution, which endows causal information
at the population level.
This level of analysis can be achieved through actual experimentation
via Randomized Control Trials, or through statistical adjustments
that smartly combine the observed conditional probabilities to 
reduce the spurious associations in the estimation. Pearl's 
$do$-calculus is likely to be the most effective approach to 
determine the identifiability of causal effects by applying the
following three rules: 1) insertion/deletion of observations, 2)
action/observation exchange, and 3) insertion/deletion of 
actions~\cite{Pearl12}.

\subsection{Language Model}
\label{subLM}

Finally, to operate with textual data, there is the need to numerically represent 
linguistic information in the former ``generic'' variables $X$. 
To this end, Probabilistic Language Models are functions that assign a 
probability to a sentence, to eventually build up a whole 
piece of text. Traditionally, in such statistical models the
sentences have been broken down, i.e., tokenized, into sequences of words, 
and the goal has been to predict the probability of an upcoming 
word~\cite{Jurafsky24,Manning99}:
\begin{equation}
\label{eqLM}
	P(X_{n+1}|X_0,...,X_{n-1},X_{n}) \ .
\end{equation}

Today, with the advent of distributed representations of words 
and phrases~\cite{Mikolov13}, along with the Transformer neural 
architecture~\cite{Vaswani17}, long texts are directly
represented in dense vector spaces,
and the task of the resulting Large Language Models is now to
provide responses to carefully engineered input 
prompts~\cite{Chen23}.

\section{Method}
\label{secMethod}

This section details the Smart Troubleshooting objectives and
the analysis procedure to attain them,
which focuses on providing root cause diagnostics and predictions
of solutions for a problem observation based on written text data.
Since the applied industrial environment belongs to the area of Predictive
Maintenance, the consideration of a common development standard such as the
ISO 13374 is recommended~\cite{iso13374}. This specification breaks down the 
complexity of a problem into small modules that may be developed 
in isolation, thus increasing the chances of project success while
also improving the interpretability and explainability of the technical
solution, and help to reduce the technical debt.
What follows is a description of the Data Manipulation and
Health Assessment processing blocks.

\subsection{Data Manipulation}
\label{subDataMan}
Causality is an emergent property of complex industrial 
systems~\cite{Yuan24}.
In this setting, linguistic variables constitute high level
qualitative descriptions that group functions into 
categories and hierarchies, as is established by the
FMMEA documentation.

\subsubsection{Return On Experience Records}
\label{subsubRoXData}
In the Smart Troubleshooting setting, the RoX text
data are collected as a means to capture and describe
the factual ontological relationships observed in the 
field~\cite{Trilla22TLP}. They display the following 
variable type structure:
\begin{itemize}
	\item Subsystem Z (common context): Categorical
	\item Root Cause C (problem source): Categorical
	\item Observation O (reported problem, failure): Text
	\item Solution S (repair/maintenance action): Text
\end{itemize}

This data structure is populated from several projects or 
environments, which exhibit some differences regarding the 
verbosity of the language used to describe the problem and 
its solution.

\subsubsection{Textual Entailment}
\label{subsubTextEntail}
The concept of entailment refers to the directional nexus 
between text fragments. Regarding the RoX data, these assumed
relations are encoded in the following graph:

\begin{figure}[h]
  \centering
\begin{tikzpicture}
\begin{scope}[every node/.style={circle,thick,draw}]
	\node[] (ss) at (1,2)  {Z}; 
	\node[] (rc) at (0,1) {C}; 
	\node[] (p) at (2,1) {O}; 
	\node[] (s) at (1,0) {S}; 
\end{scope}
\begin{scope}[>={Stealth[black]},
              every node/.style={fill=white,circle},
              every edge/.style={draw=black,very thick}]
   \path[->] 
	(ss) edge (rc)
	(ss) edge (p)
	(ss) edge (s)
	(rc) edge (p)
	(rc) edge (s)
	(p) edge (s)
        ;
\end{scope}
\end{tikzpicture}
	\caption{Graph showing the RoX variable relationships.}
	\Description{Graph showing the RoX variable relationships.}
	\label{figTextEntGraph}
\end{figure}
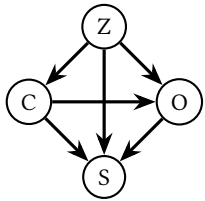

The diagram shown in Figure~\ref{figTextEntGraph} explicitly states that:
\begin{itemize}
	\item (C) is the root cause of the observed problem (O).
	\item (S) is both the effect of the observed 
		problem (O) and its root cause (C).
	\item (Z) is a general confounder, i.e., a common cause.
\end{itemize}

Once the text data for (O,S) is available in raw format it must 
be cleaned before doing any further processing. This involves
lower-casing, punctuation removal, lemmatization/stemming, 
stop word filtering, etc.

\subsubsection{Text Embeddings and Large Language Models}
\label{subsubTELLM}

Probably one of the most challenging parts of this environment is the 
embedded numerical representation of the text, which is 
typically considered as unstructured data.
The approach described in this method has been implemented
using a discrete categorical representation obtained with a
``BERTopic'' Large Language Model~\cite{Grootendorst22}. 
The proposed strategy integrates: 1)
MiniLM~\cite{Wang20}, which is a compressed version of Sentence-BERT
(i.e., a Transformer-based language model at the sentence level),
2) UMAP~\cite{McInnes18},
which reduces the dimensionality of the embedded vector space, and 3)
HDBSCAN~\cite{Malzer20}, 
which clusters and quantizes the resulting low-dimensional representation.

\subsection{Health Assessment}
\label{subHealthAss}

In the Smart Troubleshooting environment, one of the main
challenges is dealing with the confounding bias introduced by 
the diversity of subsystems and components.
To this end, Causal Inference techniques are utilized in the 
technical language processing scenario to extract relevant linguistic 
features from the text~\cite{Feder21}. 

This module exploits the probabilistic SCM, which has been
designed as a discrete category based Bayesian network following
the structure of the RoX data, and conducts a fine-grained 
diagnosis of the observed input anomaly description by determining 
its root cause, and also by providing an unbiased estimation of 
the (most likely) potential solution.

\subsubsection{Root Cause Analysis}
\label{subsubRCA}
Root Cause Analysis (RCA) is a troubleshooting method of 
problem solving used for identifying the sources of
the failures~\cite{Wilson93}.
RCA is a form of deductive inference since it requires an 
understanding of the underlying causal mechanisms for
the potential faults
and the problem, i.e., what is typically found in the context of 
Predictive Maintenance through the FMMEA documentation.

The discrete causal Bayesian Network is suitable for exploiting the 
categorized description of an observed problem (i.e., the effect)
and predicting the likelihood of the several possible known 
causes. Thus, estimating the likely root causes amounts
to computing the conditional probability diagnosis function $P(C|O)$.
Note that this estimand operates on the Observational rung of
the Hierarchy of Causality, see Section~\ref{subsubPCH1}.
Eventually, RCA yields a ordered list of potential root cause variables
along with their probabilities, which aligns with the way
complex systems fail~\cite{Cook00}. The variables that comprise 
the data are required to be representative enough to help the 
developers and engineers pinpoint the source of the observed problems
through the root causes and their effects~\cite{Weidl08}.

\subsubsection{Solution Generation}
\label{subsubPST}
Predicting the solution is especially challenging due to the large
cardinality of the Observation and Solution spaces (O,S).
To obtain an unbiased result, an (atomic) intervention shall be
performed. This represents an action $do()$ that is conducted on a system 
\emph{to set} (not filter via conditioning) its variables $X_i$ to known 
values $x_i$ and then evaluate their impact/effect on other variables 
$X_k$, i.e., $P(X_k|do(X_i=x_i))$.
This constitutes an advanced level of analysis that is not attainable with 
the observed data alone: it also needs to account for the assumptions
encoded in the causal model in the form of variable dependencies. 
As a result, the aforementioned confounding bias in the 
estimation is reduced through the following adjustment formula:
\begin{equation}
\label{eqAdjust}
	P(S|do(O)) = \sum_{C,Z} P(C|Z) \, P(S|C,Z,O) \, P(Z) \ .
\end{equation}

Note that this estimand operates on the Interventional rung of
the Hierarchy of Causality, see Section~\ref{subsubPCH2}.
Its computational burden can be somewhat alleviated if the single most
likely Cause is already determined by the former RCA procedure.

Once the representation of the most likely Solution category 
(S) is reliably determined, the 
associated text needs to be \emph{generated}. To this end,
its related textual records {\tt S} are retrieved from the
dataset and
used to \emph{prompt} a pretrained ``Llama2'' Large Language Model (LLM)
to obtain a natural language explanation~\cite{Touvron23}.

Prompt design and engineering have rapidly become essential for 
maximizing the potential and utility of a LLM~\cite{Amatriain24}.
A prompt is constructed by combining instructions, questions, 
input data, and examples. Prompt engineering requires a blend 
of domain knowledge, understanding of the AI model, and a 
methodical approach to tailor queries to different contexts.
For Smart Troubleshooting, the following query text {\tt Q} is used:

\begin{verbatim}
  Given Observation: O, with possible root Cause(s):
C, the indications for Solution used in previous 
similar cases using the predicted category are: S.
\end{verbatim}


Beyond asking a simple question, possibly the next level of 
sophistication in a prompt is to include some instructions on 
how the LLM should answer the question:

\begin{verbatim}
  You are an advanced smart troubleshooter assistant
designed to advise experts in diagnosing and solving
problems by answering questions about the possible 
solutions the expert should consider to fix the 
failure described by the Observation and Cause in 
the query. The smart troubleshooter should provide 
solutions to diagnose and solve problems. 
Additionally, the troubleshooter should provide an 
explanation for the role of each proposal and should
use appropriate forms for verbs and sentences.
  The smart troubleshooter should also refrain from 
redundancy or repetition of steps. The smart 
troubleshooter always answers as helpfully as 
possible. It is crucial that all the propositions 
should always be presented using: 
" - Option/Solution "
or any other listing format like this example layout:
" - Option 1 : here the text
  - Option 2 : here the text... "
\end{verbatim}

Additionally, the field of Causality has a priviledged position in 
developing trustworthy intelligent systems~\cite{Ganguly23}. 
For that reason, given that the pretrained LLM has learned from
a large collection of (possibly uncontrolled) documents, it is
advised to include some warning considerations
(e.g., using safe bias-free language) regarding the
integrity of the generated outcomes:

\begin{verbatim}
  The smart troubleshooter should avoid harmful, 
unethical, racist, sexist, toxic, dangerous, or 
illegal content, and ensure that the responses are
safe, socially unbiased, and genuinely positive. If
the smart troubleshooter doesn't know the answer, 
they should say so. It is crucial that the smart 
troubleshooter never provides too specific details
in their generated statements. Finally, the 
troubleshooter should follow the layout mentioned 
above for the answers and should always include any
relevant information from the Observation and 
Cause(s) given, without mentioning their indexes or
references. 

Now, give the Solution to this query: Q.
\end{verbatim}

\section{Results}
\label{secResults}

This section develops the experimental work through
one illustrative example in the Predictive Maintenance domain.
The causal model has been trained on several projects
with RoX dataset sizes between 4k and 20k records,
yielding average accuracy root cause classification scores over 80\%
(and over 70\% for precision and recall).

\subsection{Data Observation}
\label{secObs}
The specific exemplifying instance displays the following 
RoX data descriptions:

\begin{itemize}
\item {\bf Subsystem}: Suspension
\item {\bf Root Cause}: Part physically damaged
\item {\bf Observation}: ``failure mechanical brake trailer and use electrical release kph brake''
\item {\bf Solution}: ``download showed only one instance of failure on trailer with loss of comms with bcu  this fault cleared when tram was put i nto to remove brake isolation  however on tram being put back into s axle failed to apply  download showed w at fault  cor re ct reporting checked  end switch disconnected and found verdigris on pins  new loom made up and fitted  fault still present  so ch anged proximity switch  tram te ran to bulwell and back with tct  all probes alignedthrough coasting  all trailer proximity swi tches and looms cleaned and checked''
\end{itemize}

The data for this record are shown for qualitative 
comparative purposes. In a live real-world Smart Troubleshooting
setting, the maintainer or engineer shall provide the
description of the observed problem only, and the system
shall add value by producing the diagnosis
results in terms of the root cause(s) and the likely solution(s).

\subsection{Results Prediction and Retrieval}
\label{secPred}

Table~\ref{tabRCA} shows the resulting distribution 
of root causes. Note that the correct cause leads this
ranking, and the rest are given a smooth uniform value.

\begin{table}
	\caption{Ranking of the 5 most probable potential 
	root causes (out of 20 categories).}
  \label{tabRCA}
  \begin{tabular}{lc}
    \toprule
    Potential Root Cause&Probability\\
    \midrule
	  Part physically damaged & 0.9012\\
	  Accident & 0.0052\\
	  Incorrect maintenance & 0.0052\\
	  Insufficient lubrication & 0.0052\\
	  Leakage & 0.0052\\
  \bottomrule
\end{tabular}
\end{table}

Similarly, Table~\ref{tabPST} shows the resulting distribution 
of potential solutions. While the identification number of the Solution 
category is not very informative by itself, note that
the cluster that it represents is likely to have many 
descriptive instances in the dataset.

\begin{table}
	\caption{Ranking of the 5 most probable potential
	solutions (out of 302 categories).}
  \label{tabPST}
  \begin{tabular}{lc}
    \toprule
    Potential Solution&Probability\\
    \midrule
	  274 & 0.0565 \\
	  244 & 0.0169 \\
	  52 & 0.0146 \\
	  90 & 0.0135 \\
	  10 & 0.0104 \\
  \bottomrule
\end{tabular}
\end{table}

What follows is a list of the textual statements that can be 
retrieved from each of the records that share the same
leading potential solution category (i.e., the top one 
associated to the identification number 274):

\begin{itemize}
\item handover from off coming shift was to torque the hangar bolts old bolts used sourced new bolts re secured track brake torque t i ghtened and height set
\item drain oil from axle and top back up found lots of meatal on magnet when topping axle back up found that the axle has snapped bog i e needs new axle fitting
\item corroded areas addressed and all corrosion removed areas coated with two coats of loctite corrosion inhibitor
\item damaged cable removed and replaced with new
\item fitted new outer rubbers and ended up having to fit new center rubber strip to a cab as well
\item hole drilled out and re tapped bracket then fitted
\end{itemize}

\subsection{Prompt-based Generation}
\label{secGeneration}

Finally, based on the given problem Observation text, on the most 
likely Root Cause, and on the text retrieved from the most
likely unbiased Solution, the LLM can now synthesize the following 
enhanced advisory:

\begin{itemize}
\item Solution 1: Check and Replace Bolts. As the possible cause suggested is failure of handover from the off-coming shift, which resulted in torquing the hangar bolts using old bolts, it is possible that the bolts may have been loose or damaged. Therefore, check and replace the bolts to ensure they are tightened properly to secure the suspension.
\item Solution 2: Address Oil Leakage. Another possible cause mentioned is oil leakage from the axles, which can cause damage to the suspension and affect the braking system. Check for oil leakage around the axle and replace any damaged seals to prevent further damage.
\item Solution 3: Replace Snapped Earth Cable. Previous observation mentions that a snapped earth cable was found, which can cause electrical issues in the braking system. Replace the snapped earth cable to ensure proper electrical connection and prevent any further failures.
\item Solution 4: Address Corrosion. Corrosion can affect the suspension and braking system, leading to safety issues. Check for any signs of corrosion in the suspension and braking components and address them accordingly. Apply anti-corrosion measures, such as applying Loctite corrosion inhibitor, to prevent further damage.
\item Solution 5: Adjust or Replace Fixings. Previous observation mentions that some fixings were found to be below spec, which can affect the suspension. Adjust or replace any fixings that are found to be damaged or not secure enough to ensure proper suspension.
\item Solution 6: Inspect Sanding Compressor. Previous observation mentions that the sanding compressor was rebuilt, but faults were still present after testing. Inspect the compressor and its components to identify any potential issues and address them to prevent any further failures.
\item Solution 7: Address Worn Cable Insulation. Previous observation mentions that the cable insulation was found worn through, which can cause electrical issues in the braking system. Address any worn-out cable insulation to ensure proper electrical connection and prevent any further failures.
\item Solution 8: Replace Damaged Component. Previous observation mentions that a damaged mudflap was found, which can affect the suspension and braking performance. Replace any damaged components to ensure proper operation of the suspension and braking system.
\end{itemize}

It is important to note that these proposed solutions 
are based on the given information and any previous records, 
and they may not be exhaustive. However, most of them
correctly advise a replacement action (for a part that
is physically damaged).

Finally, while the way to quantitatively (i.e., objectively) 
evaluate the LLM-generated outcomes is still an open research
question driven by correlational (i.e., not necessarily causal) 
scores~\cite{Kim24},
these qualitative (i.e., subjective) results suggest a reasonably
promising future to help the subject matter experts troubleshoot
the failures in challenging industrial settings.

\section{Discussion}
\label{secDisc}

Apparently, the quality of the text generated by the LLM
seems higher than what the staff write on the RoX records: it shows
more clarity, better diction, and better spelling.
Nevertheless, there are no safety guarantees against hallucinations,
and state-of-the-art LLMs are also subject to irrational behavior
and reasoning breakdown even on simple 
tasks~\cite{MacmillanScott24,Nezhurina24}.

Up to this point, the approach presented in this workshop paper
has described the basic principles of its causal RCA and 
Solution Generation technology, and an initial experimental proof 
of concept has been shown. This early stage of maturity corresponds
to a standard ISO 16290 Technology Readiness Level (TRL) 
between 4 and 5, because it has been validated in some
real-world relevant environments~\cite{ISOTRL}.
This section brainstorms some avenues of improvement to increase
this robustness indicator up to higher quality standards,
considering the specific challenges of complex industrial environments,
and to eventually demonstrate the technology in an operational 
environment (TRL 6--7).

\subsection{Vector Database}
\label{subVecDB}
One first idea could be to improve the granularity of the
(currently discrete categorical) linguistic representation in the 
Causal Bayesian Network.
The approach presented in Section~\ref{subsubCBN} first
embeds the unstructured text data into a large vector 
space, then it reduces the dimensionality of this
real-valued numerical description, and finally it quantizes the resulting
low-dimensional representation to obtain a categorical
random variable. At each step, though, some information is
lost due to compression, and while this is especially
advantageous to decrease the complexity of the ensuing (discrete)
probabilistic model, maybe it also introduces some unnecessary 
limitations. Therefore, to potentially improve this situation, 
Hybrid Bayesian Networks may be helpful to represent 
the Observation and Solution texts with their original
vectors (note that the Subsystem and
the Root Cause variables shall retain their categorical nature).
Hybrid Bayesian approaches, which are able to simultaneously
model both discrete and continuous variables~\cite{Atienza22}, 
have already enjoyed success in multivariate
domains for predicting
delays in operations~\cite{Lessan19}, and also in the
reliability assessment of large infrastructure 
networks~\cite{Zwirglmaier24}.

In this new modeling scenario, the text may exploit the larger 
distributed representation of the LLM embedding, which is
characterized by a set of independent real-valued dimensions. For 
retrieval purposes, the most likely vector $v^*$ in the embedded 
linguistic space $V=(v_0,v_1,...)$ could first be obtained as:
\begin{equation}
\label{eqLikelyVec}
	v^* = \max_{v_0,v_1,...} \prod_{i \in V} P(v_i) \ ,
\end{equation}

and then the matching with the RoX records 
could be conducted using the cosine
distance metric that has traditionally been supported by the
statistical language processing field. However, it
remains to be seen how the curse of dimensionality will
affect the technical setting. In any case, this realignment
with the well established techniques may be of help to
increase the TRL.

\subsection{Transportability}
\label{subTransp}
Generalizing empirical findings to new environments or 
populations is necessary in the Smart Troubleshooting setting because
there are different projects and fleets considered, and each environment 
exhibits particularities in the written form of the text data.
The concept of ``transportability'' is defined as a license to 
\emph{transfer} 
information learned in one environment or domain to a different 
environment~\cite{Bareinboim13}, and thus reduce the covariate
shift problem.

Transportability analysis assumes that enough structural knowledge 
about both domains is known in order to substantiate the production 
of their respective causal diagrams.
To formally articulate this transfer procedure, a selection
variable $K$ must be introduced to represent the differences 
between the deployments. In the RoX-based Smart Troubleshooting 
setting for the industry, the assumption is that the 
only relevant difference among the environments is driven 
by the population of subsystems, thus $K \rightarrow Z$
(in fact, some components are only present in specific 
platforms and assets, so this premise is well
founded). The resulting transport formula to generate
solutions from a source environment $A$ to a target 
environment $B$ is shown as follows:

\begin{equation}
\label{eqTransport}
	P_B(S|do(O)) = \sum_{C,Z} P_A(C|Z) \, P_A(S|C,Z,O) \, P_B(Z) \ .
\end{equation}

If one particular environment $B$ is found to be especially lacking in
some aspect, then the rest of the environments $A$ can be used to
estimate the desired probabilistic distribution. This smart
workaround to a direct data shortage problem that leverages the 
indirect data from multiple settings is expected to increase 
the robustness of the predicition, which in turn
may help to increase the TRL of the final solution.

\subsection{Counterfactual Analysis}
\label{subsubPCH3}
So far, the main focus of the analysis has been put on the observed 
factual data at the population level. However, these data represent 
only one of the many potential outcomes the system could have 
experienced: had things been different, an alternative outcome
may have been observed. In this sense, a
counterfactual describes a potential 
distribution at the \emph{individual} level
driven by hypothetical speculations over data that may contradict 
the facts. This level of analysis constitutes an additional third 
rung in the Hierarchy of Causality described in Section~\ref{subCI}.
Conducting this estimation requires the following 
three steps~\cite{PearlCIS}:
\begin{enumerate}
\item {\bf Abduction}: Beliefs about the world are initially
updated by taking into account all the evidence $E$ given in the context
of a single instance/unit. Formally, the 
exogenous noise probability distributions $P(U)$ are updated to $P(U|E)$.
\item {\bf Action}: Interventions are then conducted to reflect the 
counterfactual assumptions, and a new causal model is therefore created.
\item {\bf Prediction}: Finally, counterfactual reasoning occurs over 
the new model using the updated knowledge.
\end{enumerate}

Gaining access to such involved analysis creates a new area
of research to enhance Predictive Maintenance.

\subsubsection{Algorithmic Recourse}
\label{subsubAlgoRec}

Algorithmic Recourse is an approach that systematically explores these
counterfactual worlds~\cite{Karimi22}. Such environments
are simulated via inference through (atomic) interventions 
$\alpha$ in the form of alternative problem descriptions.
This is expected to help in the recognition and understanding 
of the general root causes that lead to the system 
failure~\cite{LiRCA22}, and the solution advisory that leads to
greater availability.

Formally, the specific retrospective reasoning that these
counterfactuals explore on the reported anomaly, i.e., 
the full description of the solved problem, can be stated as:
\begin{equation}
\label{eqAlgoRec}
	P(S^*|do(O=\alpha), Z, C, O, S) \ .
\end{equation}

Given that the solution of a problem was factually implemented and
recorded in the RoX data, i.e., through observing all of the variables 
(Z,C,O,S), Equation~(\ref{eqAlgoRec}) estimates the probability 
distribution of the textual representation of the hypothetical 
Solution $S^*$ had the problem been described (and 
represented) by $\alpha$, instead of the 
numerical representation it actually had when it was written.
This sophisticated degree of surgical detail enbales driving
investigations to a deeper level, and this is regarded to help
in the advance of the TRL.

\section{Conclusion}
\label{secConc}
This workshop paper has developed a complete top-down troubleshooting
approach from first Causal Inference principles that is also 
compliant with industrial development guidelines. On the basis 
of processing technical language, the focus of this learning challenge 
has been put on creating a distributed representation of linguistic
features, and exploiting it for the purpose of obtaining unbiased
causal diagnostics and solutions. This approach has been illustrated
through a relevant example in the Predictive Maintenance domain, 
and the results arguably suggest a promising line of future 
research toward a method to evaluate generative models in 
other industrial settings.

\bibliographystyle{ACM-Reference-Format}
\bibliography{kdd24refs}

\end{document}